\documentclass[a4paper,twocolumn]{article}

\usepackage[utf8]{inputenc}
\usepackage{eurosis}
\usepackage{graphicx}
\usepackage{myHarvard}
\title{THE CHALLENGE OF BELIEVABILITY IN VIDEO GAMES:\\ DEFINITIONS, AGENTS' MODELS AND IMITATION LEARNING}
\author{
Fabien Tencé$^{*,**}$, Cédric Buche$^*$, Pierre De Loor$^*$ and Olivier Marc$^{**}$\\
$^*$ UEB -- ENIB -- LISyC\\
$^{**}$ Virtualys\\
Brest -- France\\
\{tence,buche,deloor\}@enib.fr
}
\date{}

\begin{document}

\maketitle
\thispagestyle{empty}

\begin{abstract}
In this paper, we address the problem of creating believable agents (virtual characters) in video games. We consider only one meaning of believability, ``giving the feeling of being controlled by a player'', and outline the problem of its evaluation. We present several models for agents in games which can produce believable behaviours, both from industry and research. For high level of believability, learning and especially imitation learning seems to be the way to go. We make a quick overview of different approaches to make video games' agents learn from players. To conclude we propose a two-step method to develop new models for believable agents. First we must find the criteria for believability for our application and define an evaluation method. Then the model and the learning algorithm can be designed.% 163 words
\vspace{1em}
\end{abstract}

\keywords{Adaptive decision making, believability, human-like, evaluation, imitation learning, video games.}

\section{INTRODUCTION}
\paragraph{} %VG->presence/immersion
Nowadays, more and more consoles and video games are designed to make the player feel like he/she is in the game. To define how well this goal is achieved, two criteria have been defined in academic research: \emph{immersion} and \emph{presence}. According to Slater, immersion is an objective criterion which depends on the hardware and software \cite{Slater1995}. It includes criteria based on virtual sensory information's types, variety, richness, direction and in which extend they override real ones. For example, force feedback and motion sensing controllers, surround sound and high dynamic range rendering can improve the immersion. Presence, also known as telepresence \cite{Steuer1992}, is a more subjective criterion. It is defined as the psychological sense of ``being there'' in the environment. It is mainly influenced by the content of the video game.
\paragraph{} %presence->believability
As stated in \cite{Slater1995}, presence partly depends on the match between sensory data and internal representation. This match expresses the fact that we try to use world models to better understand what we perceive and to be able to anticipate \cite{Held1991}. This idea is close to what is called \emph{believability} in the arts. Indeed, we can believe in fictional objects, places, characters and story only if they mostly fit in our models. Enhancing believability of video games' content should then enhance the presence.
\paragraph{} %believability->believable agents->learning
As there are many ways to enhance believability in video games, we choose to focus on believable virtual characters, also known as believable agents. The reason why we make this choice is because characters have often a major role in the believability of book and movie stories. However, unlike book and movie characters, agents should be able to cope with a wide range of possible situations without anyone to tell them what to do. Instead of defining manually these behavior, it can be interesting for the agents to learn them, reducing the time to design a character. The ability to learn has also the advantage of increasing believability so it should be considered as a must-be feature.
\paragraph{} %outline
The goal of this paper is to have an overview of the possibilities and constraints for the realization of a believable and learning-capable agent in a video game. We first define what are believable agents in games, how we can evaluate them and the relations between believability and the type of video games. Then, we make a quick overview of how an agent can express believable behaviors and how it can learn them. Finally, we conclude on a two-step protocol to realize believable agents.

\section{BELIEVABILITY} \label{sec_belie} %goal 

\subsection{Definition of Believability}
\paragraph{} %base definition
The notion of believability is highly complex and subjective. To define and understand this concept, we must look at its meaning in the arts where it is a factor of \emph{suspension of disbelief} \cite{Bates1992}. According to Thomas and Johnston, two core animators of Disney, believable characters' goal is to provide the \emph{illusion of life} \cite{Thomas1981}. Reidl's definition is more precise: ``\emph{Character believability refers to the numerous elements that allow a character to achieve the `illusion of life', including but not limited to personality, emotion, intentionality, and physiology and physiological movement}'' \cite[page 2]{Riedl2005}. Loyall tries to be more objective saying that such a character ``\emph{provides a convincing portrayal of the personality they [the spectators] expect or come to expect}'' \cite[page 1]{Loyall1997}. This definition is quite close to one factor of the presence, the match between players' world model and sensory data.
\paragraph{} %two definitions
If we want to apply the believability definition for video games, things become even more complex. Unlike classic arts, players can be embodied in a game by the mean of virtual bodies and can interact. The believability question is now: does a believable character have to give the illusion of life or have to give the illusion that they are controlled by a player? \cite{Livingstone2006}. There can be very important differences as even if the video game depicts the real world, all is virtual and players know that their acts have no real consequence.
\paragraph{} %which definition? why?
In this paper, we will consider only believable as ``giving the illusion of being controlled by a player''. At first glance, it can be seen as going against presence as we remind the players that there is a real world. However, not using this approach has some drawbacks too: virtual characters only giving the illusion of life can be classified as ``being a piece of program'' which may break the illusion permanently. Players may also see problems in the virtual characters' behavior only because they know they are not human-controlled.
\paragraph{} %definition->criteria
Now that we chose a definition for believability, we have to find the different criteria that influence it. Defining those criteria will make improvements and evaluation possible.

\subsection{Believability Criteria}
\paragraph{} %criteria
Criteria to define believability are highly domain-dependant. For example, for embodied conversational agents (ECA), factors like quality of the speech, facial expressions, gestures are important \cite{Ruttkay2002}. For emotional and social agents, reaction to others is the most important part \cite{Reilly1996} and so on with the numerous domain working on believable agents. However, there is one criteria that can be used for every domains. What have been named Eliza effect \cite{Weizenbaum1966} is really interesting: as long as the agent is not actively destroying the illusion, people tend to see complex thinking mechanisms where there are not.
\paragraph{} %unpredactibility but not too much
One way to destroy the illusion is predictability, which is a classic flaw for characters in video games. Characters doing over and over the exact same thing are rapidly categorized as being artificial. Adding some unpredictability can better a lot believability \cite{Bryant2006}. The difficulty is that too much unpredictability can give the feeling of randomness which can harm believability too. Another flaw is obvious failure to learn. An agent keeping the same behavior even if it is clearly counter-productive breaks the illusion of being ``intelligent''. That is the main reason why we think that believable agents must be able to learn \cite{Gorman2007}.
\paragraph{} %being able to ``see'' the behavior->agents might overdo
The last criterion we will cite makes the difference between believability and realism. Unlike realistic agents, believable agents might be forced to overdo for observers to understand what they are doing \cite{Pinchbeck2008}. Although it can seem strange, it is sometimes mandatory to exaggerate some of the agent's characteristics so that people believe in it. This technique is very often used in arts, especially in cartoons. This means that human-like agents could have a quite low believability. There are however links between realism and believability so it should be a good start to draw inspiration from realism.
\paragraph{} %criteria->eval
Knowing what influence believability, we can design an evaluation method for believable agents. As the cited criteria are quite general, more domain-specific ones should be taken into account in the evaluation.

\subsection{Evaluation of Believability}
\paragraph{} %believability evaluation: Turing's test
As believability is subjective, evaluation is a critical and complex step. Even if it was not intended to, Turing's test \cite{Turing1950} is still considered as a reference for believability evaluation. In its standard interpretation, a judge must chat with a human and a machine using only text. If, after a certain amount of time, the judge cannot tell which one is artificial, the machine is said to be intelligent. This test's goal was to assess intelligence but it has been much criticized \cite{Searle1980,Hayes1995}. This critique, however, does not apply to believability and it is even a very good basis for assessing believability as we defined earlier.
\paragraph{} %parameters in believability evaluations
There are many parameters for believability evaluation methods \cite{MacNamee2004,Gorman2006,Livingstone2006}. The first one is to cast or not to cast doubt on the nature of the presented character(s). This choice is often linked with mixing agents and humans so that the judges assess real humans' believability too. This can be useful to avoid bias induced by prejudices and to have a reference: humans usually do not score a perfect believability. Another parameter is the number of questions and answers. Turing's test features only one question and a yes/no answer whereas other tests feature many questions and scales to answer. The former choice may be too restrictive while the latter may result in too much undecided answer to ``beat'' the test. Another problem is the computation of the overall believability score which, in case of multiple questions, may give experimenters too much influence on the results. To add more objectivity, it is possible to have relative scoring instead of absolute: the score given to a example can answer to ``is example A better than example B?''. It is also necessary to decide if judges are players or only spectators. While players can actively test evaluated characters, spectators are more focused on them and can notice much more details in the behaviors. Finally the choice of the judges is really important. Cultural origins \cite{MacNamee2004} and level of experience \cite{Bossard2009} may have a noticeable impact on believability scores.

\subsection{Links between Believability and Video Games}
\paragraph{} %Multiplayer
Following our definition of believability, multiple players should be able to play together in the same video game. Indeed, players must not know \textit{a priori} if what they perceive is due to another player or to a program. Believable agents should be used to replace a missing player, having the same role as the others so that we can achieve high level of believability. Of course, single player games can achieve human-like agents but they will not be believable as we defined it.
\paragraph{} %Interactions between players
The video games' complexity of interactions has a really important impact on the believability. If interactions between the players, agents and the game are few, it may be hard to assess believability. For example in a pong game, players can only move paddles so it may be difficult for them to guess the thoughts of the adversary. On the contrary, if players and agents have a wide range of possible actions, they can better perceive the others' thoughts making believability harder to achieve. Video games featuring a complex world and allowing a lot of interactions should be preferred to develop and evaluate believable agents.
\paragraph{}%type
Video games are often categorized in different groups. For a testbed, we need a multiplayer game with a lot of possible interactions between players. This kind of games can be found in the action, role playing, adventure and sport games categories. From a technical point of view, adventure and sport games tend to be difficult to modify and in particular to add agents. Role playing involves a large part of communication and natural language is far too much complex to make conversational agents truly believable. Action games are often good choice because they have a quite limited set of actions, simplifying the learning, but their combination are complex enough to be a challenging problem for believability.
\paragraph{} %FPS
The classical example of such action games is first person shooter games. In those games, each player or agent controls a unique virtual body and sees through its eyes. The character can, non-thoroughly , grab items (weapons, \dots{}), move (walk, jump, \dots{}) and shoot with a weapon. Each character has an amount of hit points, also known as life points: each time an actor is hit by an enemy fire, a certain amount of hit points are subtracted to the current value. When hit points reach zero, the character ``dies'' but can usually reappear at another place in the game's environment. Although the concept can seem very basic, it can prove challenging to design believable agents. To make things simpler at the beginning, it is possible to avoid cooperative rules as tactics used may be quite hard to learn.

\section{AGENTS' BEHAVIOUR MODELLING} \label{sec_model} %how to express our solution
\paragraph{} %perception/decision/action
The general principle of an agent is to perceive its surroundings, take some decisions according to those perceptions and its internal state and then make some actions which depend on the decisions. This is named the perception, decision, action loop because the sequence is repeated over and over. The sets of actions and perceptions are defined by the video game so designing an agent is ``only'' to find a good mapping between perceptions and actions.
\paragraph{} %what is a model
The general structure of the agents' thinking mechanisms is called a model. The model, when parametrized, generates behaviors: it gives a sequence of actions when fed with a sequence of perceptions. In other words, the model is the way to express behavior and the parameters are the behavior.
\paragraph{} %many models
There are many different agents' behavior models and some can achieve better believability than others. Three domains are very influential on their design: computer sciences, psychology and neurosciences. Computer sciences see agents' models as programs, with inputs and outputs and composed of mathematical functions, loops, etc. Models influenced by psychology try to apply what science understood about humans' behaviors. Finally, models based on neurosciences try to code an artificial brain composed of neurons grouped in different specialized zones. There are many hybrid models and even some of them combines all the three.
% \paragraph{}%connectionism/cognitivism
\paragraph{} %reactive vs deliberative
There are also two different approaches in the way the agents think: reactive or deliberative. Reactive agents map directly actions to perceptions. The problem is that they tend not to act with long term objectives. Deliberative agents, on the other hand, try to predict the results of their actions. They make plans so that the sequence of their actions will satisfy their needs. Their main weakness is that they are often not very good in rapidly changing environments. Some agents combine reactive and deliberative abilities to have both advantages without the weaknesses.

\subsection{Models in Industry}
\paragraph{} %believable models in industry (FSM, etc)
In the video game industry, models are often quite basic but can achieve good believability in the hands of skilled game designers. Finite state machines (FSM), hierarchical FSM and behaviors trees are widely used. Those models can express logic, basic planning and are good at reactive behaviours. Another advantage is that those models are deterministic, in other words we can predict what can happen and then try to avoid problems. Their major weakness is that their expressiveness is limited and explaining complex behaviors can become unmanageable.
\paragraph{} %believable models in industry (planners)
Another type of model which is used in video game industry is planners. They can give very good results in terms of believability, an example is the game F.E.A.R. which is partially based on the STRIPS planner \cite{Fikes1971}. Agents featuring planners are able to set up complex tactics which have an important impact on believability. However they are not yet capable of learning so they can become predictable. Other models are, by less, used in video games.

\subsection{Models in Research}
\paragraph{} %believable models in research not suitable
In scientific research, many different models are used for agents. Some of the main architectures are Soar \cite{Newell1987}, ACT-R \cite{Anderson1993} and BDI \cite{Rao1995}. However, these models are not design for believability but for reproducing some humans' thinking mechanisms. Some models are, from the beginning, designed for believability like those used in \cite{Lester1997,MacNamee2004} and in the Oz project \cite{Bates1992,Reilly1996,Loyall1997} but they are not suited for action games but for emotions and social interactions.
\paragraph{} %for action games
For action games, some work has been done and different models have been presented. \cite{LeHy2004} is a Bayesian model which seems to produce quite believable behaviors but lacks of long-term goals. \cite{Robert2005} is a model using classifiers with a online unsupervised learning algorithm but the agents' believability has not been evaluated. \cite{Gorman2006} use Bayesian imitation, partly based on \cite{Rao2004}, and have a good believability. However it only produces a part of the whole behavior needed for the game. \cite{Gorman2007} is ``only'' based on neural networks but due to its learning algorithm, it gives really good results in terms of believability. Here too, it produces only a part of the behavior.
\paragraph{} %choices
A good choice seems to have an hybrid reactive-deliberative agent as planning has a great impact on believability so as reactivity \cite{Laird2000}. Contrary to what is commonly used in video games, a non-deterministic approach is preferable because it can increase believability by avoiding repetitiveness. Agents must be able to learn and, in the best case scenario, should need minimal \textit{a priori} knowledge. This implies that the model does not need to be easily understandable and adjustable as most of its knowledge could come from learning.

\subsection{Actions and Perceptions}
\paragraph{} %general choice
The choice of the model's inputs and outputs is very important because it has a great influence on the final result. What seems logical is to have the exact same sets for the agents and for the players so that agents have all the necessary tools to be believable. However, the perceptions and actions information follow a complicated process between the players' brain and the game. Moreover, having human-like interaction can make the problem even more complicated and instead of improving believability, it could make it worse.
\paragraph{} %perceptions
The perception information begins in the video game, each object in it having a set of characteristics like position, rotation, color or texture. All those objects are rendered so they are a sequence of pixel. Those pixels are displayed by a peripheral like a screen or a video projector. Then the information go to our eye and then to our brain. We use visual information as an example, it is also true with sounds, touch and will be true for odors and tastes in the future.
\paragraph{} %perception point of view
As we can see, the perception information takes multiple forms and all may not be usable or useful for believable agents. From the moment the game's environment is rendered, we enter the domain of artificial vision, audition, etc. Unfortunately, this is a very difficult problem and very few information can be obtained this way. As a consequence, there is a great loss of available information when the rendering is done. The best compromise is to use information directly from the game. It results in having both very precise information, like position, and unavailable information, like lighting.
\paragraph{} %action
For the actions, they begin in our brain, then muscles, go through the game's peripherals and arrive in the game. It is possible to simulate peripherals' activity, however it has some drawbacks too. Peripheral information are very low-level so we must translate high-level goals to low-level sequences. If the agent wants to go to some place, it has to find a sequence of peripheral-like actions instead of defining a trajectory for its virtual body.

\section{LEARNING BY IMITATION} \label{sec_learn} %how to reach our goal
\subsection{Why Imitation?}
\paragraph{}
Form the beginning, we have discussed only about learning without specifying the ``how''. The reason we choose to focus on imitation only is because of believability. As we defined it, believability means ``to look like a player''. One efficient way to achieve this is to copy or imitate him/her \cite{Schaal1999}.
\subsection{Definition and Approaches}
\paragraph{} %definition
Imitation has quite a pejorative connotation as the ``innovate, don't imitate'' sentence shows. It is however a clever way of learning and very few animals can use it \cite{Blackmore1999}. Moreover, humans use it very often, such often that they do not always notice it and they do it from the very beginning of their lives \cite{Meltzoff1977}. Imitation is the act of observing a certain behavior and roughly repeating it. This repetition can be deferred and the demonstrator may or may not be aware of being imitated.
\paragraph{} %behavioral and cognitive sciences approach
From a behavioral and cognitive sciences approach, imitation is the increased tendency to execute a previously demonstrated behavior. The imitated behavior must be new to the imitator and the task goal should be the same as the demonstrator \cite{Schaal1999}. There are some debates on whether or not the strategy should be the same. According to Meltzoff, inferring intentions is necessary for higher level imitation where imitators learn the goals instead of the strategy to achieve them \cite{Rao2004}. This can lead to unsuccessful attempts being use as examples in imitation learning as long as the goal can be inferred. It this case, the strategy may be different from the example given by the demonstrator.
\paragraph{} %neuro approach
From a neuroscience approach, imitation can be explained studying the brains of evolved animals. Some areas are hypothesized to be involved in imitation learning according to studies on macaques' brains. The quite recent discovery of \emph{mirror neurons} may help in understanding the mechanisms of imitation. Mirror neurons are neurons that fire both when we do something and when we look at someone doing the same thing. However it is not sure that they actually have a major role in the imitation process.
\paragraph{} %computer sc approach
In the domains of robotics and artificial intelligence, imitation learning becomes more and more popular. It has been particularly used in movements imitation in robotics because it reduce the programming time of humanoids robots \cite{Gaussier1998}. Indeed, when we explain a movement to somebody else, showing the example is far more productive than explaining only with words. The same goes with robots and agents: imitation reduce the space of hypothesis the program has to search.

\subsection{Imitation Learning in Computer Sciences}
\paragraph{} %interest (believability)
The main reason we chose imitation learning is that believability is quite linked with resembling to the demonstrator, in our case humans. According to some authors, imitation can lead to believable or at least humanoid agents and robots \cite{Schaal1999,Thurau2005}. This learning technique seems to be both fast and adequate for our goal.
\paragraph{} %state of the art in AI (proba)
An interesting model and learning algorithm based on imitation has been designed by Rao and Meltzoff \cite{Rao2004}. What makes it particularly interesting is that it is based on Meltzoff's work on imitation in infants. The model uses probabilities to determine which actions should be taken according to the perceptions. Those probabilities are learnt observing a player's virtual body. Another interesting characteristic is that the agent can guess the player's goals to better understand his/her behavior. This model and algorithm have been extended and used in a action game \cite{Gorman2006} leading to interesting results.
\paragraph{} %state of the art in AI (neural gas, force field)
Some work focused more on the imitation of movement in games which is often highly reactive and strategic at the same time. In their work, Thurau \textit{et al.} used an algorithm named neural gas to represent the geometry of the game's environment \cite{Thurau2004a}. They then use an algorithm to learn potential for each neuron/waypoint by giving path followed by players. The sum of those potentials form a field force, attracting the agent to interesting places.
\paragraph{} %state of the art in AI (neural net)
Another interesting work does not use a new model or a new algorithm \cite{Gorman2007} and it only covers one behavioral aspect: aiming and shooting in an action game. The model consists of three neural networks, one for choosing a weapon, one for aiming and one to choose to fire or not. Those neural networks are trained with a Levenberg-Marquardt algorithm on players' data previously gathered and treated. The results are very encouraging as the agents were capable of short-term anticipation and acted in a really human-like fashion. Agents even copied behaviors due to the use of a mouse: right-handed players have more difficulties to follow targets travelling from the right to the left with their cursor.
\paragraph{}%lehy
One last work based on imitation learning which seems to give very good results is \cite{LeHy2004}. For basic motion control it use the rule of succession to find probabilities of action by counting the number of occurrences. This rule's main interest is that it estimates the probabilities of actions that did not occur. For the linking of the tasks to achieve, as the model is close to a hidden markov model (HMM), an incremental Baum-Welch algorithm is used to find the parameters. This algorithm permits online learning so it may prove to be a great tool for believability. The main drawback is that we must find a way to identify the current task so that corresponding parameters can be modified.

\subsection{Learning Constrains}
\paragraph{} %constraints (imitation is fast process, we want online learning, model)
As we said earlier, obvious failure to learn can break the illusion of life. Our first constraint is that the learning process can be observable. This implies that the agent can learn while playing and that it learns fast. The first implication is named \emph{online learning} in terms of artificial intelligence. It means that the model is modified for each given example. It contrasts with \emph{offline learning} where the model is only modified when all examples are given.
\paragraph{} %fast learning
Despite the fact that imitation learning can be a fast way to learn, this goal is not easy to achieve. The difficulty in imitation learning is generalization: we do not want the agent to be a film of what the demonstrator showed but to adapt to the environment as the demonstrator did. Learning systems often need examples and counter-examples to learn to avoid over-generalization and over-learning. In imitation learning we only have examples so we might need many of these for the agent to understand how to generalize. This can slow down a lot the learning.
\paragraph{} %model
As we saw earlier, learning algorithms and agents' models are heavily linked together. Indeed, the process of learning consists in modifying the model's parameters. The algorithm can even modify the structure of the model. The choice of the learning algorithm or its design cannot be done without choosing or knowing the behavioral model. This makes this choice even more complicated as we must consider the pros and the cons for the couple and not for each one separately.

\section{CONCLUSION} \label{sec_conclu}
\paragraph{} %goal
In this paper we outlined the problem of creating believable agents in video games. This problem has three parameters: the players, the video game and the agents. We can add on top the learning mechanisms to make agents capable of evolution. All those factors have an important impact on the final result so their influence should be carefully studied.
\paragraph{} %believability
First, believability is a very complex notion because of its subjectivity. We defined it as the ``illusion that the agent is controlled by a player''. Although this definition is quite different from the one used in arts, it makes sense in video games. Indeed, several people can interact in the same game and replacing some of them by computer programs can be very interesting. What makes believability complex is that although players have difficulties in understanding the others' behaviors and often interpret them wrongly, some small flaws can destroy the whole illusion.
\paragraph{} %VR system
The next important factor is the video game itself. A believable agent in an game where it is only possible to chat will be very different from a believable agent in an game where it is only possible to move. Complex games which require both reactive and planning abilities as well as featuring a wide range of possible interactions between players are the most challenging testbeds for researchers.
\paragraph{} %models
The agent itself and more precisely the model defining its behaviors is the most obvious factor in the problem. Models can express more or less believable behaviors depending on how they model human characteristics. So, although there are many models both in the industry and the research, they are not all suitable for believable agents.
\paragraph{} %imitation learning
An agent will not be able to make the illusion last very long if it is not capable of evolving. One possible way of evolution is learning. Imitation learning seems to fit perfectly with the goal of believability. Indeed, it consists in copying the demonstrator acts which is quite close to the notion of believability. It is also a very fast method for transmitting knowledge from one person to another.
\paragraph{} %future: evaluation+VE
The first step in the design of a believable agent is to define an evaluation method. This will help us in knowing which characteristics are important for our agent. This cannot be done without choosing a video game first because the notion of believability depends on the agent's set of actions. It can be also very interesting to evaluate humans' and simple agents' believability first to better understand the problem.
\paragraph{} %future: model+learning
The next step is to choose or design a behavioral model and an imitation learning algorithm. This should be done according to the believability factors discovered in the previous step. Considering the constrains of fast and online learning which make it noticeable, this reduces greatly the available models and algorithms usually used in artificial intelligence.

\section{ACKNOWLEDGMENT}
This study has been carried out with the financial support from Région Bretagne.

\bibliographystyle{myBibStyle}
\bibliography{library}

\end{document}